\documentclass[a4paper, 10pt, conference]{ieeeconf}      

\IEEEoverridecommandlockouts                              

\usepackage{graphicx}
\graphicspath{{figs/}}
\usepackage{amsmath}
\usepackage{amssymb}
\usepackage{algorithm}
\usepackage{algpseudocode}
\usepackage{times}
\usepackage{bm}
\usepackage{color}
\usepackage{mathtools}
\usepackage{mathrsfs}
\usepackage{float}
\usepackage[maxfloats=1024]{morefloats}
\newcommand{\figref}[1]{Fig.~\ref{#1}}
\newcommand{\tabref}[1]{Table~\ref{#1}}

\newcommand{\SO}{\mathop{\mathrm{SO}}}
\newcommand{\SE}{\mathop{\mathrm{SE}}}

\newcommand{\Pos}{\mathop{\mathrm{Pos}}}
\newcommand{\Rot}{\mathop{\mathrm{Rot}}}

\renewcommand{\S}{\mathbb{S}}
\newcommand{\x}{\boldsymbol{x}}
\newcommand{\y}{\boldsymbol{y}}
\renewcommand{\u}{\boldsymbol{u}}
\newcommand{\q}{\boldsymbol{q}}

\newcommand{\R}{\mathbb{R}}

\newcommand{\secref}[1]{Section~\ref{#1}}
\newcommand{\enuref}[1]{\ref{#1})}

\newcommand{\normal}{\mathcal{N}}
\newcommand{\vmises}{\scalebox{0.6}{$\boldsymbol{\mathcal{V}}$}\hspace{-0.2ex}\mathcal{M}}
\newcommand{\bingham}{\mathcal{B}}

\newcommand{\diag}{\mathop{\mathrm{diag}}}
\newcommand{\eigval}{\boldsymbol{\lambda}}

\newcommand{\rlink}{\boldsymbol{p}}
\newcommand{\uinput}{\u}
\newcommand{\qlink}{\q}

\newcommand{\idx}[1]{{\langle#1\rangle}}

\renewcommand{\mid}{\hspace{0.2ex}|\hspace{0.2ex}}

\newcommand{\tref}{\bm{\theta}^{\text{ref}}}
\newcommand{\teff}{\bm{\theta}^{\text{eff}}}

\usepackage{pifont}

\definecolor{mygreen}{rgb}{0.0, 0.75, 0.0}

\title{\LARGE \bf
Online Estimation of Self-Body Deflection With Various Sensor Data Based on Directional Statistics 
}

\author{Hiroya Sato$^\text{1}$, Kento Kawaharazuka$^\text{1}$, Tasuku Makabe$^\text{1}$,
Kei Okada$^\text{1}$ and Masayuki Inaba$^\text{1}$
\thanks{This work has been submitted to the IEEE for possible publication. Copyright may be transferred without notice, after which this version may no longer be accessible.}%
\thanks{$^\text{1}$Authors are with Department of Mechano-Informatics, Graduate School
of Information Science and Technology, The University of Tokyo, 7-3-1
Hongo, Bunkyo-ku, Tokyo, 113-8656, Japan.
{\tt\footnotesize [h-sato, kawaharazuka, makabe, k-okada, inaba]@jsk.t.u-tokyo.ac.jp}
}%
}

\begin{document}

\maketitle
\thispagestyle{empty}
\pagestyle{empty}

\begin{abstract}
In this paper, we propose a method for online estimation of the robot's posture.
Our method uses von Mises and Bingham distributions as probability distributions of joint angles and 3D orientation, which are used in directional statistics.
We constructed a particle filter using these distributions and configured a system to estimate the robot's posture from various sensor information (e.g., joint encoders, IMU sensors, and cameras).
Furthermore, unlike tangent space approximations, these distributions can handle global features and represent sensor characteristics as observation noises.
As an application, we show that the yaw drift of a 6-axis IMU sensor can be represented probabilistically to prevent adverse effects on attitude estimation.
For the estimation, we used an approximate model that assumes the actual robot posture can be reproduced by correcting the joint angles of a rigid body model.
In the experiment part, we tested the estimator's effectiveness by examining that the joint angles generated with the approximate model can be estimated using the link pose of the same model.
We then applied the estimator to the actual robot and confirmed that the gripper position could be estimated, thereby verifying the validity of the approximate model in our situation.

\end{abstract}

\section{INTRODUCTION}

In robot task realization, alignment of the body to the target position and posture is fundamental.
It requires that the posture of the actual robot approximates well that of the geometric model.
However, to achieve a highly accurate positioning, frequent maintenance of robots is required since the robot's positional accuracy deteriorates, for example, due to aging. 
These costs are barriers to using robots in environments where frequent maintenance is difficult, such as living environments.
In such environments, a robot system is needed that does not require the geometric model and actual robot posture to be very close.

If the pose difference between the actual robot and the geometric model is allowed, it is necessary to estimate how much the posture of the actual robot differs from the model. In this paper, we propose a method for online estimation of the actual robot's posture using available sensor information and propose a robot system that uses the estimator. 

For the sequential estimation of states, it is common to use Bayesian filters.
Much research has been conducted on the probabilistic approach to estimating a robot's posture. They are mainly based on visual information from the outer camera. In \cite{GarciaCifuentes2017}, 
they proposed the method for estimating self-posture from RGB-D images, considering the latency of yielding depth images.
In \cite{Tao2020}, they construct a mapping from the robot's observable features to the joint angle using the fixed camera to estimate the robot's joint angles. 
Instead of visual information, a method that estimates joint angles from IMU sensors is proposed in \cite{Cantelli2015}, in which the extended Kalman filter (EKF) was adopted. 
In \cite{von2016human}, the method for estimating human pose from a camera and IMU sensors is provided.
More generally, the method to fuse multimodal sensors is proposed
in \cite{Lee2020}, which uses a neural network to learn the dynamics model.

In general, estimating a pose, especially the orientation, is difficult because the orientation lies in non-Euclidean space.
One way to apply a stochastic filter on such space is to consider the tangent space at a point in the space so that it can be locally assumed as Euclidean space and apply a well-known filter theory, such as the Kalman filter \cite{Brookshire2016, Lu2022}. This approach can be considered a generalization of EKF.
Although the tangent space approximation considers only the neighborhood of a certain point, it works well because filtering updates state locally.

\begin{figure}[t]
    \centering
    \includegraphics[width=\linewidth]{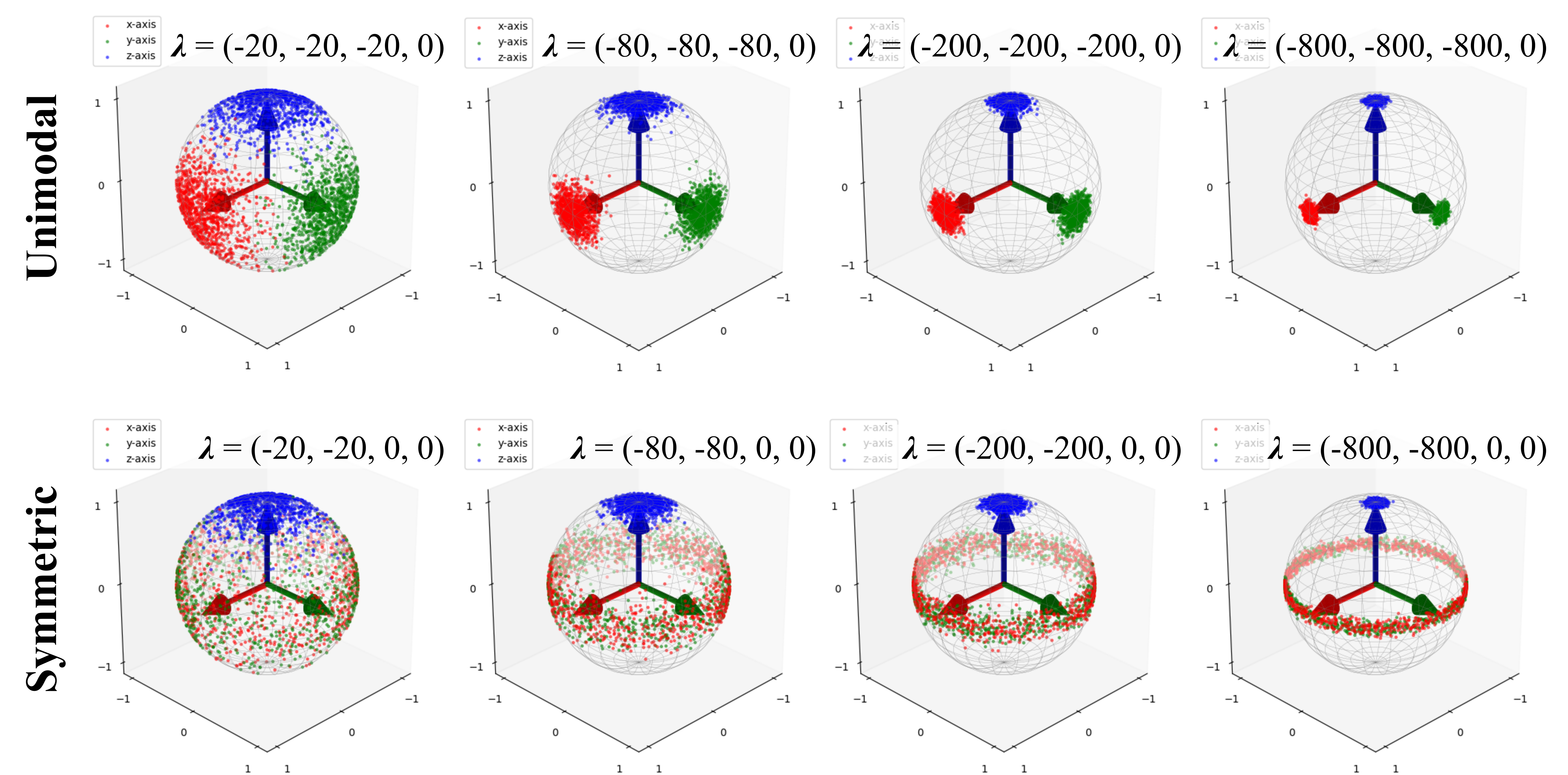}
    \caption{
        Example of Bingham distribution $\bingham(M, \eigval)$, where $M = I_4$. This is an example of the directional statistical distribution used in this paper.
    The corresponding $\eigval$ is shown in each figure. 
    Details are described in \secref{section:binghamdistribution}
    }
    \label{fig:binghamexample}
  \end{figure}

Another approach is to adopt probability distributions using in directional statistics, such as von Mises distribution and Bingham distribution.
These have been introduced into the fields of sensor fusion.
A von Mises filter is used for a system containing circular variables in its state \cite{chen2015, CHEN2016543}, achieving higher accuracy than the traditional EKF.
Bingham distribution has also been employed for recursive filtering \cite{igor2014, binghamfilter2018}.
They have used the distribution mainly to achieve high-performance filtering or overcome the discontinuity of quaternion parametrization \cite{Zhou2019OnTC}, so the distribution often has only a single mode (unimodal; see \figref{fig:binghamexample}).
Unlike the tangent space approximation, these distributions can handle a global property of probability density all over the space.


In our method, we adopt directional distributions as noise distributions for the state variable.
In particular, we apply them to the observed noise to obtain a probabilistic representation of the global characteristics of the sensor.
As an example of its application, this paper deals with the yaw drift of an IMU sensor using not only a unimodal Bingham distribution but also an axis-symmetric distribution (symmetric; see \figref{fig:binghamexample}).

The detailed results are presented in \secref{section:experiments}.
In addition, our method uses a particle filter for the following reasons.
First, it can easily handle nonlinear observation noise via the likelihood function.
Second, attitude estimation is often multimodal, i.e., observations can be reproduced in multiple ways, so it is generally unsuitable to use the method which assumes a target as a single distribution, such as the unscented Kalman filter (UKF).

\section{PROBABILITY DISTRIBUTIONS}

\subsection{Multivariate Gaussian distribution}

We define the $m$-variate Gaussian distribution as follows:
\begin{equation}
\begin{aligned}
  &\normal^m(\mu, \Sigma)(x) \\ 
  =&\frac{1}{\sqrt{(2 \pi)^m \det{\Sigma}}} \exp \left(-\frac{1}{2}(x-\mu)^{\top} \Sigma^{-1}(x-\mu)\right),
\end{aligned}
\end{equation}
where $x \in \R^m,\,\mu \in \R^m$, and $\Sigma \in \R^{m\times m}$ is a symmetric positive definite matrix.
If a random variable $X$ has the distribution $\normal^m(\mu, \Sigma)$, we write
\begin{equation}
  X \sim \normal^m(\mu, \Sigma).
\end{equation}
This notation ``$\sim$'' will be used for other distributions.

\subsection{von Mises distribution}

\begin{figure}[t]
  \centering
  \includegraphics[width=\linewidth]{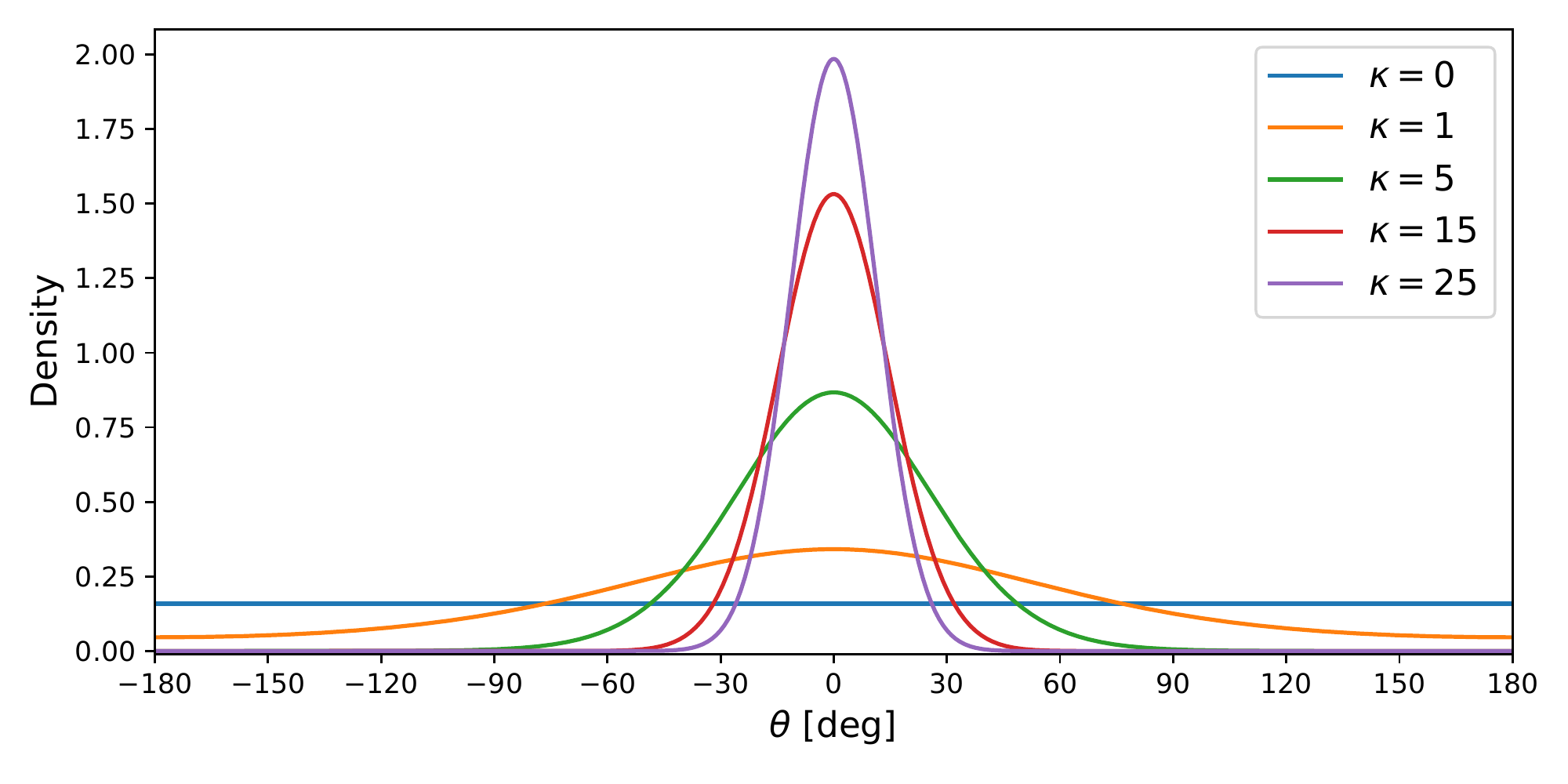}
  \caption{Some examples of von Mises distribution. Note that the left ($\theta = -180$) and right ends ($\theta = +180$) of this figure are connected and looped.}
  \label{fig:vonmisesexample}
\end{figure}

\begin{figure}[t]
  \centering
  \includegraphics[width=\linewidth]{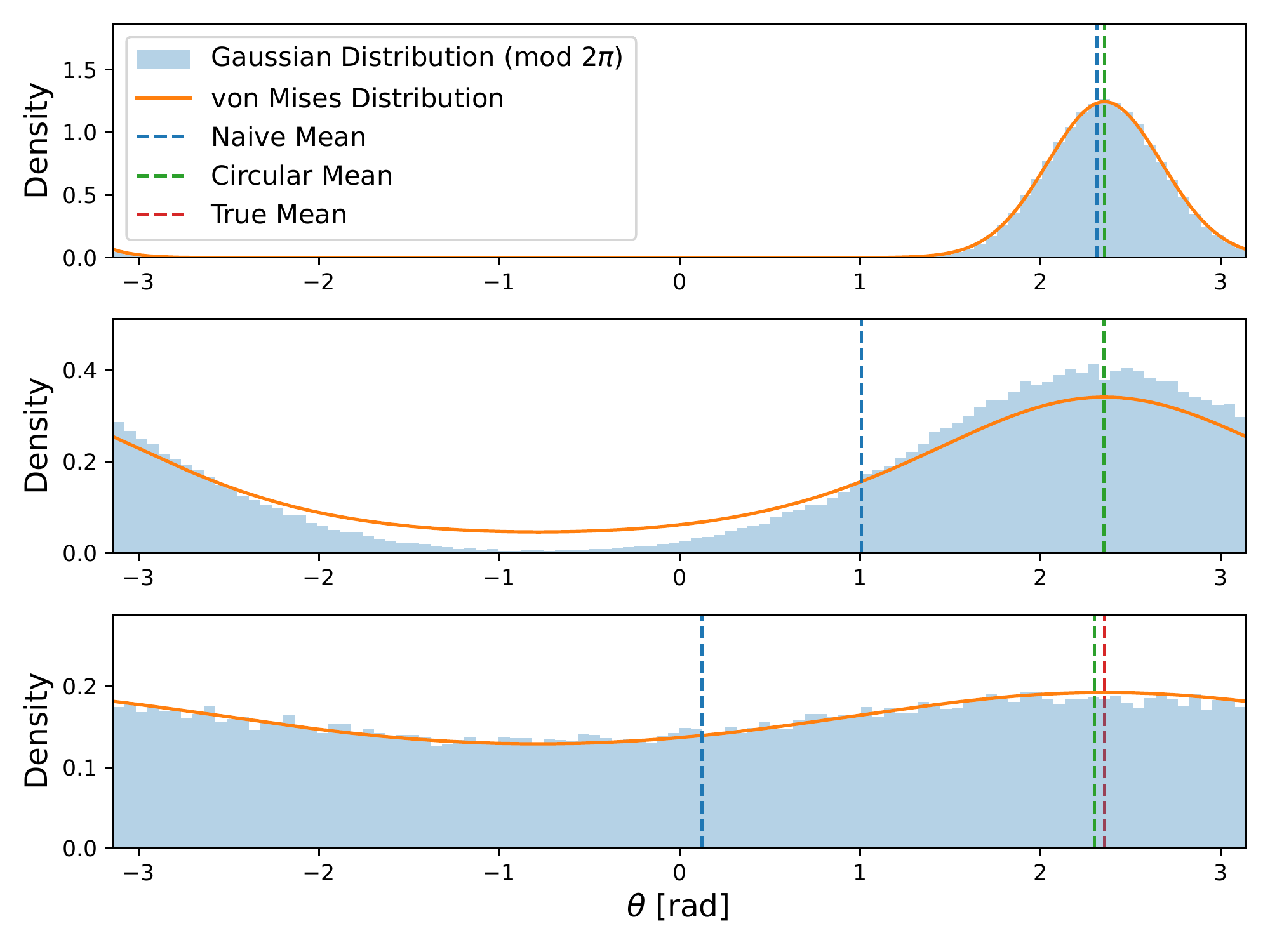}
  \caption{Comparison between von Mises distribution and Gaussian distribution.
  ``Gaussian distribution'' shown in the figure is the histogram of the points $(x \mod 2 \pi) - \pi$, where $x \sim \normal(\theta_0, 1/\sqrt{\kappa}), \theta_0 = 3\pi/4$.
  From the top, $\kappa = 10.0,\, 1.0,\, 0.20$ respectively.}
  \label{fig:vonmises_comparison}
\end{figure}


We define the von Mises distribution as follows:
\begin{equation}
  \vmises(\kappa, \theta_0)(x) = \frac{1}{2 \pi I_{0}(\kappa)} \exp \left(\kappa \cos (x-\theta_0)\right),
\end{equation}
where $x \in [-\pi, \pi), \, \theta_0 \in [-\pi, \pi),\, \kappa \in [0, \infty)$. $I_{0}(\cdot)$ is the modified Bessel function of 
the first kind of order zero \cite{Mardia1999}.
$\kappa$ is the concentration parameter, and the distribution becomes dense if $\kappa$ is large and vice versa.

\figref{fig:vonmisesexample} shows some examples of the von Mises distribution. Note that this distribution is cyclic: $\vmises(\kappa, \theta_0)(x + 2n\pi) = \vmises(\kappa, \theta_0)(x)$ for every integer $n$.
\figref{fig:vonmises_comparison} shows the density function of von Mises distribution and the histogram of sampled points from Gaussian distribution.
For large $\kappa$, it is known that the following holds 
\cite{Mardia1999}.
\begin{equation}
  \vmises(\kappa, \theta_0) \approx \normal\left(\theta_0, \frac{1}{\sqrt{\kappa}}\right)
\end{equation}
Crucially, one cannot estimate the mean of $\theta$ by simply calculating the arithmetic mean of $\theta$, which is represented in \figref{fig:vonmises_comparison} as ``Naive Mean''. Instead, we must calculate
\cite{Hassan2012}:
\begin{equation}
  \bar{\theta} = \mathop{\mathrm{atan2}}(\sin(\theta), \cos(\theta)). \label{eq:circluarmean}
\end{equation}
This $\bar{\theta}$ is shown as ``Circular Mean'' in \figref{fig:vonmises_comparison}. 

Since our system is based on the particle filter, we can choose the von Mises distribution as the system distribution instead of Gaussian. In this paper, we adopt von Mises distribution for system distribution.
We will get a closer look in \secref{sec:hyperparameters}

If $x_i \sim \vmises(\kappa_i, \theta_{0,i})$ independently for $i=1,\dots,m$ and $x = (x_1, \dots, x_m)^\top $, we write
\begin{equation}
  x \sim \overbrace{\vmises(\kappa_1, \theta_{0,1})(x_1)\times \dots \times \vmises(\kappa_m, \theta_{0,m})(x_m)}^{m}.
\end{equation}

\subsection{Bingham distribution}
\label{section:binghamdistribution}


The $m$-dimensional Bingham distribution is an antipodally symmetric distribution over a unit sphere $\S^m = \{\x\in \R^{m+1} \mid \x^\top \x = 1\}$, and defined as follows
\cite{bingham1974}:
\begin{equation}
  \bingham^m(D, \eigval)(\x) = \frac{1}{\mathcal{C}(\eigval)} \exp\left( \x^\top D \diag(\eigval) D^\top \x \right), \label{eq:binghamdefinition}
\end{equation}
  where $\x \in \S^m, \eigval \in \R^{m+1}$, and $D \in \R^{(m+1)\times (m+1)}$ is an orthogonal matrix. $\diag(\eigval)$ is diagonal matrix whose diagonal components are $\eigval$. $\mathcal{C}(\eigval)$ is the normalizing constant. Crucially, if we set $\boldsymbol{1}_m = {\underbrace{(1,\dots,1)}_m}^\top$, and $c\in \R$, we get
  \begin{equation}
    \bingham^m(D, \eigval + c\boldsymbol{1}_m)(\x) = \bingham^m(D, \eigval)(\x). \label{eq:binghamshift}
  \end{equation}
  Therefore, we can shift $\eigval$ to set one of the components to zero.

  The Bingham distribution is suitable to represent the distribution over unit quaternions due to its antipodally symmetric property.
  We write the set of unit quaternions also as $\S^3$. 
  We can set 
  \begin{equation}
    \eigval = (\lambda_1, \lambda_2, \lambda_3, 0)^\top, \label{eq:eigval}
  \end{equation} 
  where $\lambda_1 \leq \lambda_2 \leq \lambda_3 \leq 0$, thanks to \eqref{eq:binghamshift}.
  If we set $\eigval$ as \eqref{eq:eigval}, it is convenient to adopt ``xyzw'' notation for quaternion instead of ``wxyz'', because the mode of the distribution becomes the identity quaternion when we set $M = I_4$.
  Here $I_m$ denotes the $m$-dimensional identity matrix.
  Throughout this paper, we adopt ``xyzw'' notation: that is, if we have a quaternion $q = w + xi + yj + zk$, where $i,j,k$ are imaginary units and $x,y,z\in\R$,
  we identify it with $(x,y,z,w)^\top \in \S^3 \subset \R^4$.

  \figref{fig:binghamexample} shows the distribution with various $\eigval$s. One can see that if the magnitude of the component in $\lambda$ becomes larger, then the distribution becomes more concentrated.
  The Bingham distribution can represent an axis-symmetric distribution, as shown in the bottom row of \figref{fig:binghamexample}, which plays an important role in attitude estimation using IMU sensors.
  The estimation will be described in 
  \secref{section:IMU}.

\section{METHODOLOGY}

\subsection{Particle Filter}

We abbreviate $\{\x_i \,|\, i=1,\dots,t \}$ to $\x_{1:t}$.
The same applies to $\y_{1:t}$.
The system and observation equations are defined as follows:
\begin{align}
  \x_t &= \boldsymbol{f}_t(\x_{t-1}, v_t), \label{eq:xt_t-1} \\
  \y_t &= \boldsymbol{h}_t(\x_{t}, w_t), \label{eq:obsvyt}
\end{align}
where $v_t$ is a system noise and $w_t$ is a observation noise.
This implies holding the Markov model as defined below.
\begin{equation}
  \x_{t} \sim p(\x_t | \x_{t-1}), \quad
  \y_{t} \sim p(\y_t | \x_{t}).
\end{equation}
Here the following Markov property holds.
\begin{align}
  p(\x_{t} | \x_{1:t-1}, \y_{1:t-1}) &= p(\x_t | \x_{t-1}), \\
  p(\y_{t} | \x_{1:t}, \y_{1:t-1}) &= p(\y_t | \x_t).
\end{align}
Eq. \eqref{eq:obsvyt} requires that every observation $\y_t$ can be written using $\x_t$ explicitly.
Under these assumptions, the sequential filtering can be written as follows \cite{Gordon1993}.
\begin{align}
    p(\x_{t} \mid \y_{1: t-1}) &=\int p(\x_{t} \mid \x_{t-1}) p(\x_{t-1} \mid \y_{1: t-1}) d \x_{t-1} ,\\
  p(\x_{t} \mid \y_{1: t}) &\propto p(\y_{t} \mid \x_{t}) p(\x_{t} \mid \y_{1: t-1}).
\end{align}

\begin{figure}[t]
  \centering
  \includegraphics[width=\linewidth]{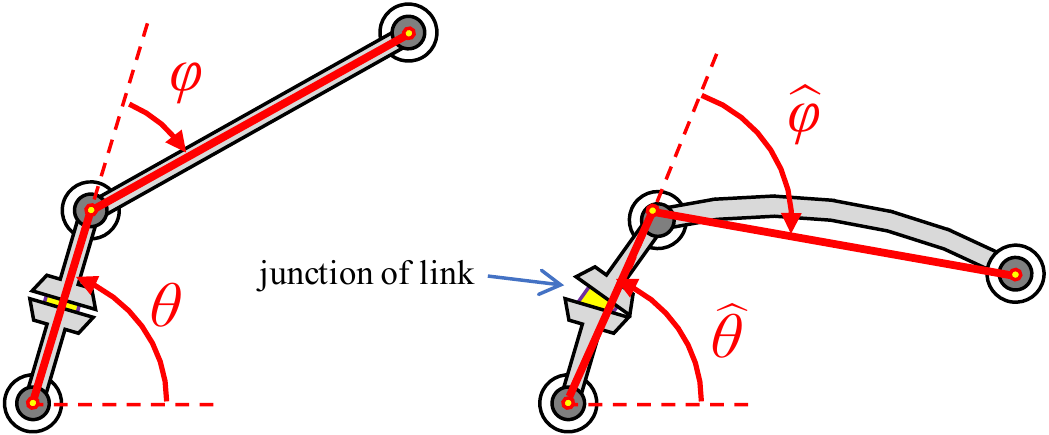}
  \caption{A diagram for explanation of an effective joint angle. We assume that the deflection of links or joints can be represented by adjusting their angles. In our situation, $\uinput_t = (\theta, \varphi)^\top$ and $\x_t = (\widehat{\theta}, \widehat{\varphi})^\top$.}
  \label{figure:approx-qhat}
\end{figure}



In particle filtering, 
the distribution $p(\x_{t-1}\mid \y_{1:t-1})$ is approximated by \textit{particles} $\{\x_{t-1\mid t-1}^{(i)}\}_{i=1}^N$ as follows.
\begin{equation}
  p(\x_{t-1} \mid \y_{1:t-1}) \approx \frac{1}{N} \sum^N_{i=1} \delta(\x_{t-1} - \x_{t-1\mid t-1}^{(i)}),
\end{equation}
where $N$ is the number of particles. Applying \eqref{eq:xt_t-1}, 
the filtered distribution can be calculated as:
\begin{align}
  p(\x_t \mid \y_{1:t}) &= \sum^N_{i=1} \beta_t^{(i)} \delta(\x_t - \boldsymbol{f}(\x_{t-1\mid t-1}^{(i)}, v_t)),\\
  \text{where}\quad \beta_t^{(i)} &= \frac{\ell_{t|t-1}^{(i)}(\y_t)}{\sum_{i=1}^N \ell_{t|t-1}^{(i)}(\y_t)}.
\end{align}
Here we abbreviate
\begin{equation}
  \ell^{(i)}_{t|t-1}(\cdot) = p(\cdot \mid \x_{t\mid t-1}^{(i)}).
\end{equation}
$\ell_{t|t-1}^{(i)}(\y_t)$ is usually called a \textit{likelihood function} of $\y_t$.
In practice, it becomes too small to handle directly because of underflow. Instead of direct calculation, it is stable to compute $\log (\ell_{t|t-1}^{(i)}(\y_t))$ first, and then calculate
\begin{equation}
  \beta_t^{(i)} = \frac{\exp[ \log (\ell_{t|t-1}^{(i)}(\y_t)) - \log (\ell_{t|t-1}^{(M)}(\y_t))]}{\sum_{i=1}^N \exp[ \log (\ell_{t|t-1}^{(i)}(\y_t)) - \log (\ell_{t|t-1}^{(M)}(\y_t))]},
\end{equation}
where $M = \arg \max_{i=1}^N \log (\ell_{t|t-1}^{(i)}(\y_t))$.
In resampling, we used the stratified resampling method \cite{Li2015}.




\subsection{Assumption in our Estimation}
\label{section:modelassumption}
In our method, we assume that the posture of the actual robot can be reproduced by correcting the joint angles of the rigid model (we will use the words ``geometric model'' and ``kinematics model'' in the same meaning).
In other words, as in \figref{figure:approx-qhat}, we consider that any differences in posture from the rigid model caused by deflection of links or rattling of joints of the actual robot can be absorbed by adjusting the joint angles.

Under this assumption, we take the joint angles of the rigid body model as the state vector. 
We predict the positions and orientations of each link by calculating from the rigid body model and then modify only the joint angles to be consistent with those of the corresponding links estimated from observed sensor information.
Here we call this joint angle the \textit{effective joint angle} and write it as $\teff$. We write the \textit{reference joint angle} as $\tref$. Using this notation, we can abstractly formulate our settings as follows.
\begin{equation}
  \mathop{\text{Posture}}\left(\tref ; \text{Actual}\right) = \mathop{\text{Posture}}\left(\teff ; \text{Rigid}\right),
\end{equation}
where the first entry of $\mathop{\text{Posture}}$ is a joint angle, and the second is a robot we are considering.

\subsection{Likelihood Functions for Directional Statistical Distributions}

In the context of filtering, for notational simplicity, $\tref$ is rewritten as $\uinput_t$ and $\teff$ as $\bm{x}_t$.
Letting $\y_{t}$ be a tuple $\y_{t} = (\uinput_t, \rlink_t, \qlink_t)$, 
the likelihood function $\ell_{t|t-1}^{(i)}(\y_t)$ can be written as follows.
\begin{equation}
  \ell_{t|t-1}^{(i)}(\y_t) = \ell_{t|t-1}^{(i)}(\uinput_t) \cdot \ell_{t|t-1}^{(i)}(\rlink_t)
  \cdot \ell_{t|t-1}^{(i)}(\qlink_t).
  \label{eq:likhood}
\end{equation}
Here we assume that $\uinput_t$, $\rlink_t$, and $\qlink_t$ are conditional independent given $\x_{t\mid t-1}^{(i)}$.

Before describing each component in \eqref{eq:likhood}, we define some notation.
$\Rot: \SE(3)\to \S^3$ denotes a function that returns the quaternion 
of the input's rotation part.
Note that the quaternion $q$ and $-q$ represent the same rotation.
To make $\Rot$ well-defined, we choose the output quaternion with the non-negative scalar part.
Similarly,
$\Pos: \SE(3)\to \R^3$ denotes a function that returns the position of the given 6D pose.
Written in mathematical formula, for $T \in \SE(3)$ satisfying $T(\x) = R\x + \bm{t}\,\,(R\in \SO(3),\, \bm{t} \in \R^3)$ for all $\x \in \R^3$, then
\begin{equation}
  \Pos(T) = \bm{t},\quad \Rot(T) = q,
\end{equation}
where $q = (x,y,z,w)^\top \in \S^3$ is the quaternion that satisfies $w\geq 0$ and the following equation.
\begin{equation}
  R = \left(\hspace{-1.5mm}
  \begin{array}{ccc}
      1-2y^{2}-2z^{2} & -2 w z+2 x y & 2 w y+2 x z \\
      2 w z+2 x y & 1-2x^{2}-2z^{2} & -2 w x+2 y z \\
      -2 w y+2 x z & 2 w x+2 y z & 1-2x^{2}-2y^{2}
  \end{array}\hspace{-1.5mm}\right)\hspace{-1mm}.
  \label{eq:quaternion2so3}
\end{equation}

We write the pose of the frame at given joint angles $\x_t \in [-\pi, \pi)^m$ as $\{F; \x_t\} \in \SE(3)$.
Let $\mathscr{F}_P$ (resp. $\mathscr{F}_Q$) be a set of the links whose position (resp. orientation) is to be observed. 
For example, if we choose links $\{EE\}$ to observe position and $\{L_1\}, \{L_2\}$ to observe rotation, then
$\mathscr{F}_P = \{\{EE\}\},\,\mathscr{F}_Q = \{\{L_1\}, \{L_2\}\}$.
In this notation, likelihood functions of $\rlink_t$ and $\qlink_t$ are defined as follow.
\begin{align}
  \ell_{t|t-1}^{(i)}(\rlink_t) &= \hspace{-1.5ex}\prod_{\{F\} \in \mathscr{F}_P} \hspace{-1.5ex}\normal^3(0,R_w)(\rlink_t^{\{F\}} - \Pos(\{F; \x_{t\mid t-1}^{(i)}\})), \\
  \ell_{t|t-1}^{(i)}(\qlink_t) &= \hspace{-1.5ex} \prod_{\{F\} \in \mathscr{F}_Q} \hspace{-1.5ex} \bingham^3(I_4, \eigval_w)(\qlink_t^{\{F\}} \odot \Rot(\{F; \x_{t\mid t-1}^{(i)}\})^{-1}),
\end{align}
where $\rlink_t^{\{F\}}$ and $\qlink_t^{\{F\}}$ are the position and the orientation, respectively, of the frame $\{F\}$.
Here $\odot$ denotes the product of quaternions.
If $\mathscr{F}_P = \emptyset$, we define $\ell_{t|t-1}^{(i)}(\rlink_t) = 1$. Similarly, if $\mathscr{F}_Q = \emptyset$ then $\ell_{t|t-1}^{(i)}(\qlink_t) = 1$.

We assume that the joint angles of the actual robot are close to the reference joint angles $\uinput_t$. To reflect this assumption, we define the likelihood function of $\uinput_t$ as follows.
\begin{equation}
  \ell_{t|t-1}^{(i)}(\uinput_t) = \prod_{j=1}^m \vmises(\kappa_w, 0)(\uinput_t\idx{j} - \x_{t\mid t-1}^{(i)}\idx{j}),
  \label{eq:kappaw}
\end{equation}
where $\uinput_t\idx{j}$ and $\x_{t\mid t-1}^{(i)}\idx{j}$ denotes the $j$-th joint's reference angle and estimated angle, respectively.







\begin{figure}[t]
  \centering
  \includegraphics[width=\linewidth]{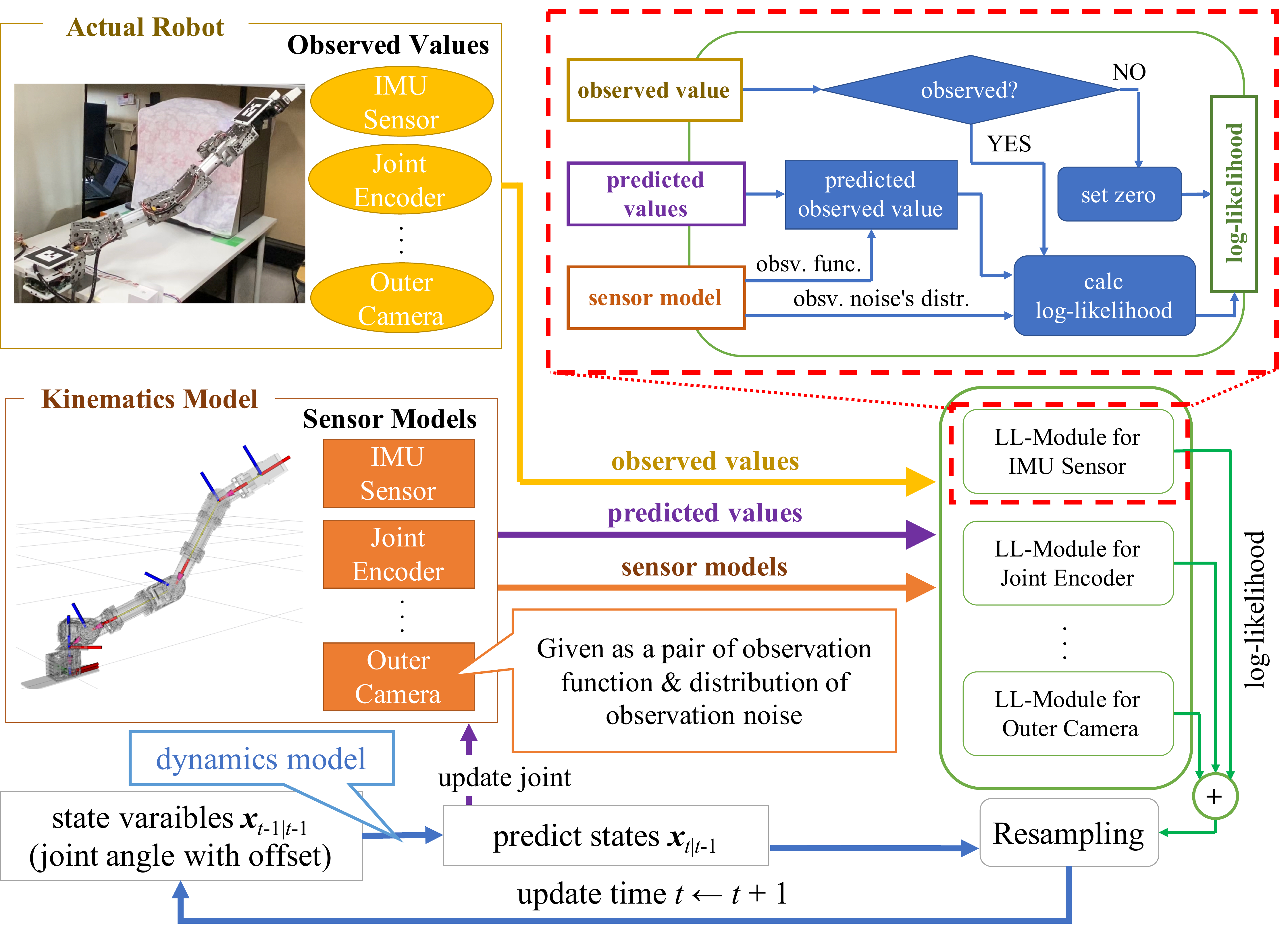}
  \caption{Overview of our system. Each color corresponds to the each values or parameters. The area surrounded by the red dashed line in the upper right corner of the figure is a detailed explanation of the components of LL-Module. ``obsv.'' = observed, ``func.'' = function, ``distr.'' = distribution.}
  \label{fig:wholesystem}
\end{figure}

\subsection{Likelihood-Based Pose Estimation System}

\figref{fig:wholesystem} shows the overview of our system.
The loop at the bottom of the image corresponds to particle filtering.
The initial particles $\x_{0|0}^{(i)} \in [-\pi, \pi)^m$ are generated for $i = 1,\dots,N$, that obey the following distribution.
\begin{equation}
  \x_{0|0}^{(i)} \sim \overbrace{\vmises(\kappa_0, 0)\times \dots \times \vmises(\kappa_0, 0)}^{m} = \vmises(\kappa_0, 0)^m.
\end{equation}
Here $\kappa_0$ is a common constant. We set $\kappa_0 = 0.1$ here. We empirically found that the filtering succeeded for any $\kappa_0 < 10.0$.
One can sample them using such as Scipy's \verb|scipy.stats.vonmises|.

With the dynamics model, a particle of one-step evolved joint angles $\x_{t|t-1}^{(i)}$ (which in our setup corresponds to an effective joint angle) can be calculated. We call these particles ``predicted particles''.
The posture of the rigid robot model corresponding to each particle $\x_{t|t-1}^{(i)}$  is calculated with the kinematics model.
For each of these calculations, the pose of each link $\{F; \x_{t|t-1}^{(i)}\}$, where $\{F\} \in \mathscr{F}_P \cup \mathscr{F}_Q$, is computed.
These are used as ``predicted values'' to compare with the values observed from the actual sensors. 
When matching the predictions with the observed values, information about the observed noise is also needed, which is given as a ``sensor model'' together with the ``observation function''.

The part of the likelihood function required for particle resampling is shown as an ``LL-Module''. The details are depicted in the area enclosed by the red dashed line.
The observation function is needed to reshape the predicted values if they do not directly correspond to the observed sensor values.
There is a conditional branch here depending on whether the sensor information is observable.
If sensor information is unavailable (e.g., hidden marker), then the corresponding log-likelihood is set to zero.
In this way, the estimator is prevented from stopping its operation if some sensors are not observed.

\section{EXPERIMENTS} \label{section:experiments}

  \begin{figure*}[t!]
    \centering
    \includegraphics[width=\linewidth]{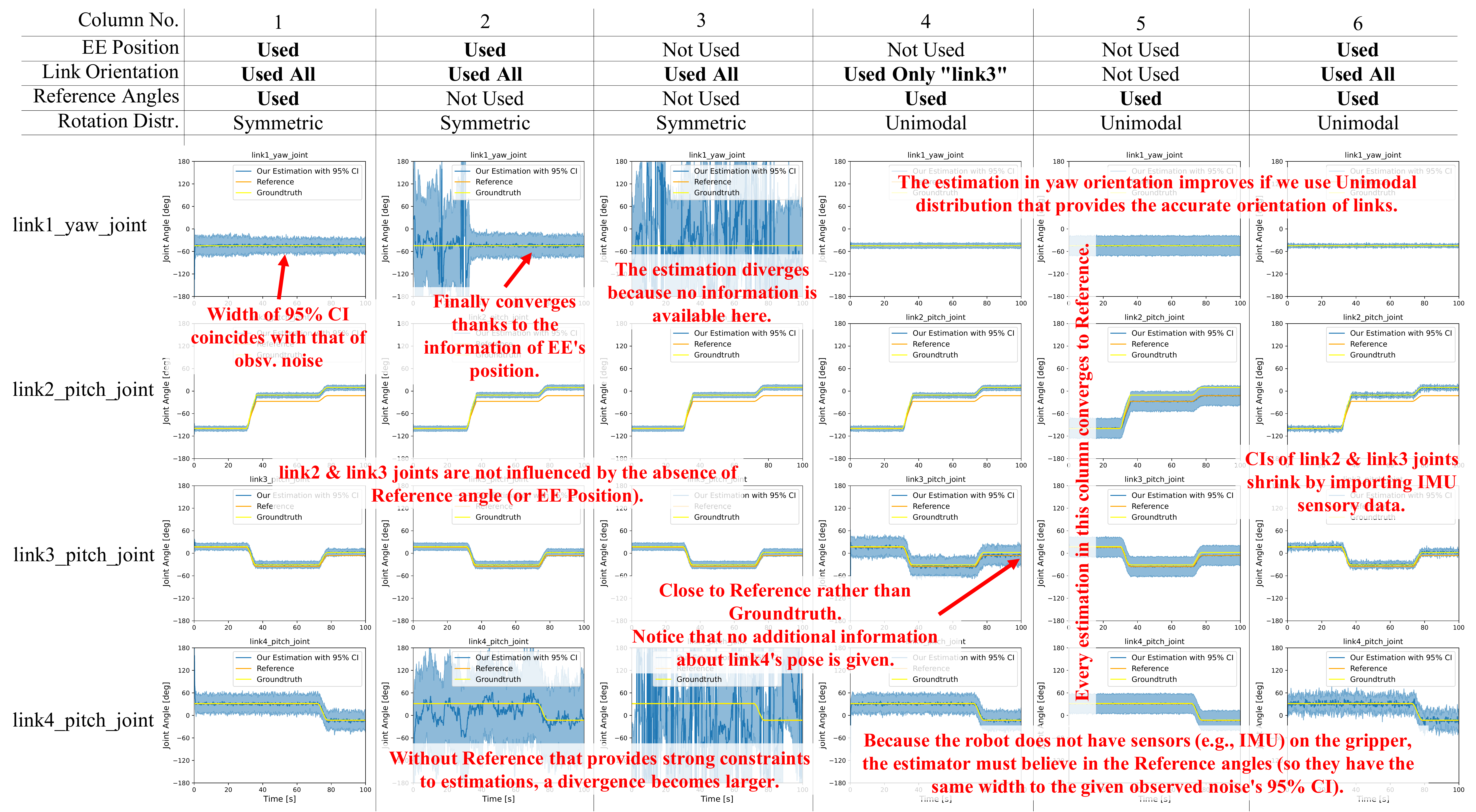}
    \vspace{-6mm}
    \caption{Some fusion results using our method with simulation. The legend and the limits of $x$-axis and $y$-axis are the same for each figure. The robot's reference postures are shown in \figref{fig:sim_posture}. CI = confidence interval, EE = end effector.}
    \label{fig:simulationresult}
    \vspace{-5mm}
  \end{figure*}

\subsection{Our Settings}

In our experiments, we assume that the joint states are constant, i.e., quasi-static motion. 
In other words, using the notation in \eqref{eq:xt_t-1},
\begin{equation}
    \boldsymbol{f}(\x_{t}, v_t) = \x_{t} + \boldsymbol{v}_t
\end{equation}
where $\boldsymbol{v}_t \sim \vmises(\kappa_v, 0)^m$. 
The value of $\kappa_v$ we set in our experiments is shown in \tabref{tab:hyperparameters}.

\subsection{Simulation Study}

\subsubsection{Simulating bent body}
We simulate the body deflection by adding offset angles to the joint reference angles.
This corresponds to the case where our assumptions described in \secref{section:modelassumption} are guaranteed to be correct.
If we set the generalized gravity at reference joint angles
$\tref$ as $g( \tref )$,
effective joint angles are calculated as follows 
\begin{equation}
    \teff = \tref + k \cdot g(\tref)
\end{equation}
where $k$ is a proportional constant. 
Considering that the resulting posture is not too close to the reference, we set $k=0.033$ here. The resulting posture is more bent than our actual robot. 
In our implementation, the generalized gravity torque is calculated by Pinocchio \cite{pinocchioweb}.

\subsubsection{Simulation Fusion result}

\begin{figure}[t]
    \centering
    \includegraphics[width=\linewidth]{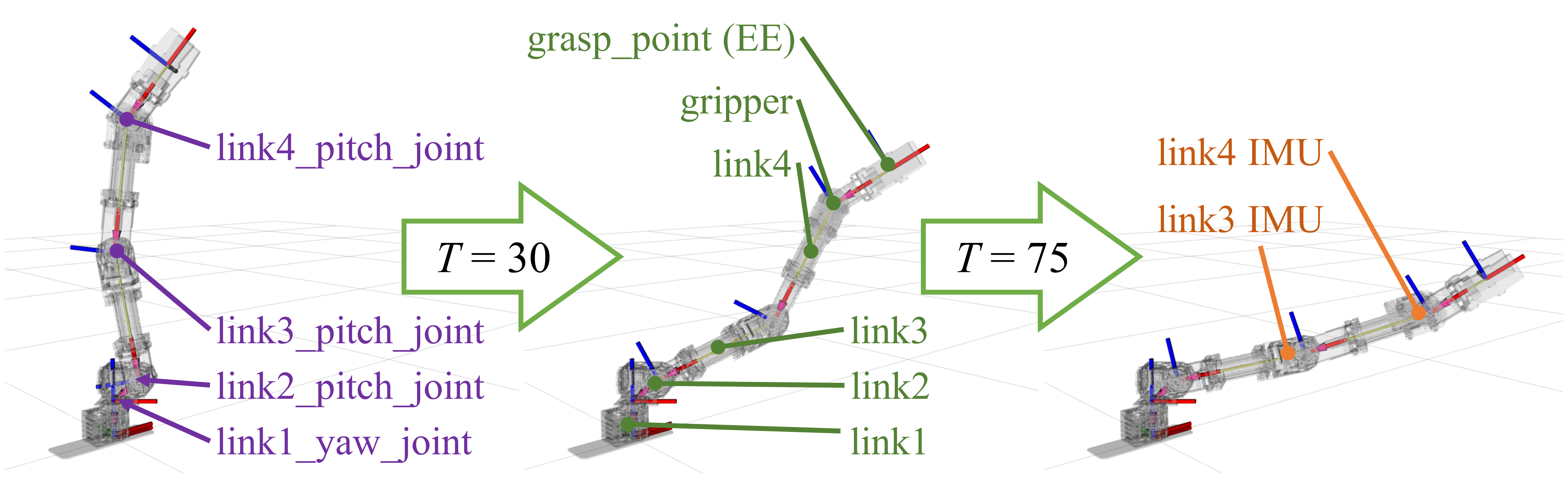}
    \caption{The transition of the reference posture in the simulation shown in \figref{fig:simulationresult}. Joint names, link names, and the placement of sensors are also shown.}
    \label{fig:sim_posture}
    \vspace{-6mm}
\end{figure}

We fuse the virtual sensor data in various combinations to see how the estimation changes with the given data.
\figref{fig:simulationresult} shows some examples of combinations. 
\figref{fig:sim_posture} shows the posture corresponding to the reference angle shown in \figref{fig:simulationresult}.
Specific discussion for each result is shown in the figure with red text.
In this simulation, we set the parameter as shown in \tabref{tab:hyperparameters}. These parameters are common with actual robot experiments (except \verb|grasp_point|, which is not used for estimation but for groundtruth of the gripper position).

\begin{figure*}[t]
    \centering
    \includegraphics[width=\linewidth]{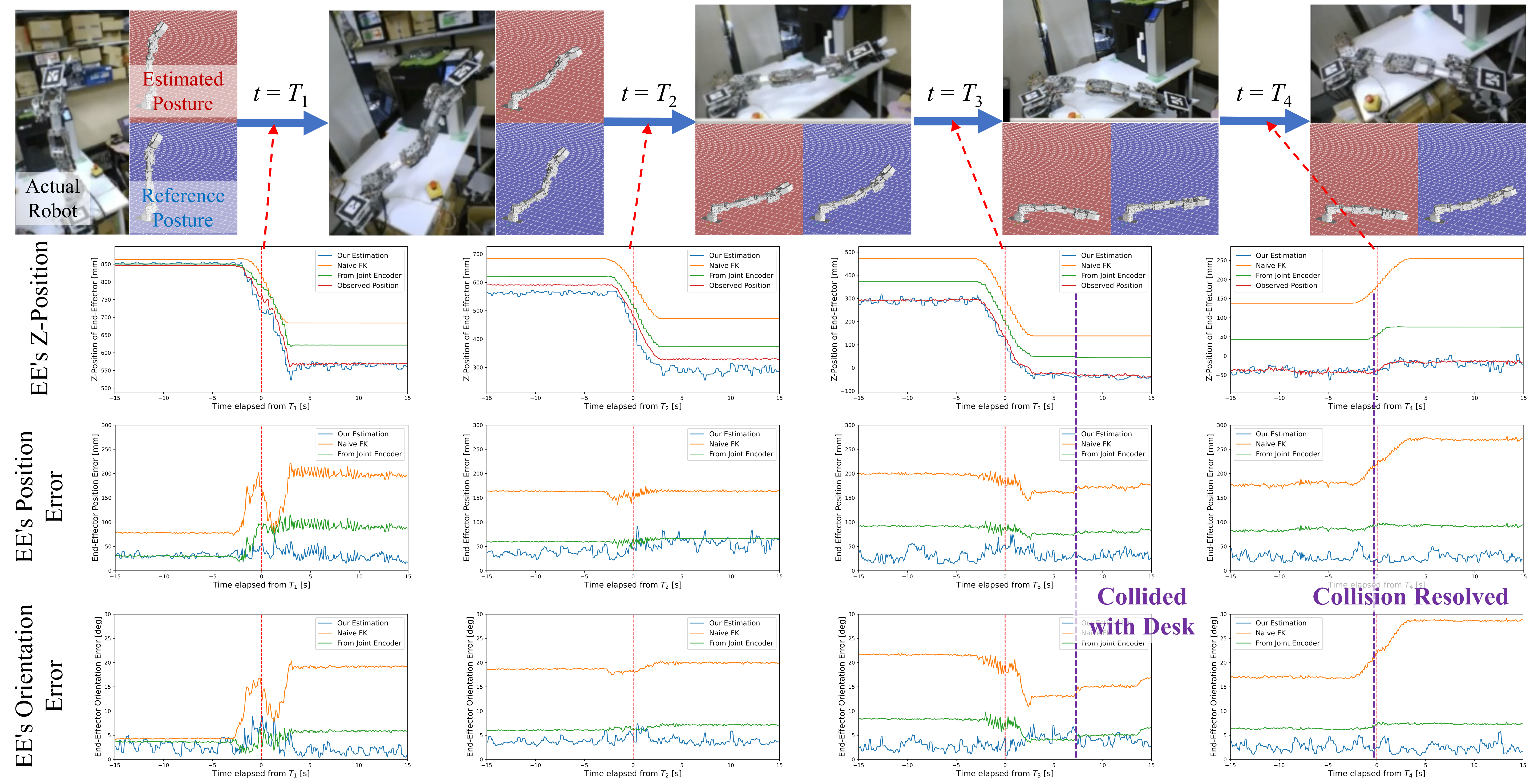}
    \vspace{-3mm}
    \caption{Result of the filtering experiments. The figures in the top row are the robot postures with the estimated joint angles (upper row) and the joint reference angles (lower row). The middle and bottom rows describe the error of the end effector (EE) from the observed position (the middle row) and orientation (the bottom row). In each figure, blue lines represent our results, which are the closest to the zero lines for each posture. This implies that our assumption still hold for the actual robot.}
    \label{fig:results}
  \end{figure*}

    

\subsection{Estimating Gripper Position From Estimated Effective Joint Angles}



Next, we check that our filter also works for cases where our assumptions do not necessarily hold.
We confirmed the validity of our assumption by comparing the estimated effective joint angles with the actual gripper positions observed from an external camera.
We put the AR markers on the robot's hand and base link for giving the reference.

\figref{fig:results} shows the estimation result with the actual robot. 
The corresponding posture is indicated at the top of the image.
When the robot's posture is vertical and not affected much by gravity, the gripper position estimated directly from the joint encoders shows good approximation performance.
As the robot's posture approaches horizontal, the effect of gravity-induced deflection increases, and the position error estimated directly from the geometric model (Naive FK) also increases.
Importantly, the accuracy of the estimation was better for the values incorporating the IMU sensor than for the values estimated only from the joint encoder.
This suggests that our approximation method, which corrects only the joint angles, is valid in our problem setting, even for the actual robot.

During the experiment, there was a situation where the robot accidentally collided with the desk, which is shown in \figref{fig:results}.
Even at this moment, our estimates work well, suggesting that they are robust to disturbances. Further investigation of the robustness to external disturbances is the subject of future work.

\begin{table}
    \centering
    \caption{Hyperparameters for estimation for the actual robot. Each distribution are origin-centered: $M = I_4$ if Bingham, $\mu = 0$ if Gaussian, $\theta_0 = 0$ if von Mises.}
    \begin{tabular}[t]{c|c|c}
        name & distribution & value \\ \hline
        \verb|link2_pitch_joint| & Bingham & $\eigval = [-800, -800, 0, 0]$ \\
        \verb|link3_pitch_joint| & Bingham & $\eigval = [-800, -800, 0, 0]$ \\
        \verb|grasp_point| (EE) & Gaussian & $\Sigma = \diag([0.01, 0.01, 0.01])$ \\
        reference joint angles & von Mises & $\kappa_w = 15.0$ \\
        system noise & von Mises & $\kappa_v = 5.0$ \\
        number of particles & - & $N=2048$
    \end{tabular}
    \label{tab:hyperparameters}
\end{table}

\subsection{Estimating Effective Joint Angles from IMU}

Intuitively, we might expect that we can extract the effective joint angles by calculating the difference between the orientation of the joint's parent and child links.
Here we compute the estimated effective joint angle by replacing the joint angles calculated from the IMU measurement for the second and third components of the joint encoder's measurement.
The value from the joint encoder is used in the first and fourth components: the first is used because the 6-axis IMU sensor cannot measure the yaw direction without drift, and the fourth is used because the gripper is not equipped with an IMU sensor.
We called this process a \textit{direct fusion}. We set this as a baseline of the fusion result.

The IMU sensor data are filtered by applying Madgwick filter \cite{madgwick2010efficient} without magnetic option.
In the direct fusion, we processed as follows:
\begin{enumerate}
    \item Yield the estimated orientation from the IMU sensor by filtering.
    \item Calculate the appropriate Euler parameter for each orientation to extract the yaw component.\label{enum:appropriate}
    \item Set the yaw rotation to $0$ in order to cancel the drift.
    \item Calculate back quaternions from the yaw-canceled parameters.
    \item Calculate the difference of angles of two quaternions and adopt the result as ``effective joint angle''. \label{enum:difference}
\end{enumerate}
In \enuref{enum:appropriate}, we have two orders of the Euler parameter to extract the yaw component: 'xyz' and 'yxz'. Due to the gimbal lock, the estimation will fail if we choose inappropriate order of parameters.
In \enuref{enum:difference}, we compute the angle of \texttt{link2\_pitch\_joint} by calculating the difference between the IMU sensor result and the identity quaternion. 

\begin{figure}[t]
    \centering
    \includegraphics[width=\linewidth]{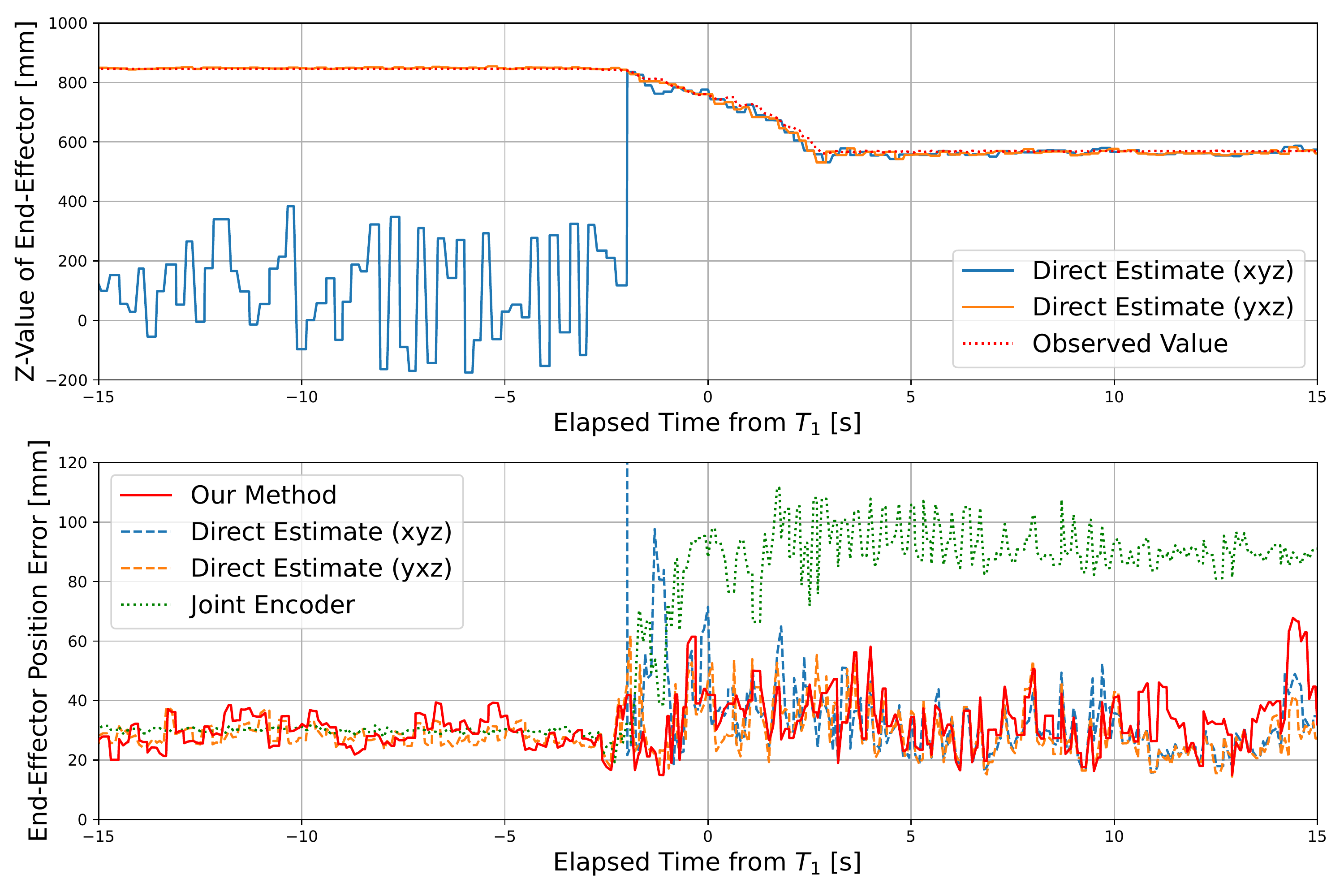}
    \caption{The results of direct fusion from IMU sensors on the actual robot. The top rows show $z$ components of the end-effector position estimated with direct fusion. The bottom row shows the reference joint angles sent to the robot. The fusion collapses if the order of the Euler parameter is not carefully chosen due to the gimbal lock (see $T < 0$). In $T>0$, the direct fusion and our result yield similar values (the middle row), which implies that our fuser can import IMU sensor data correctly.}
    \label{fig:directestim}
\end{figure}

\figref{fig:directestim} shows the result of the direct fusion and our method.
In direct estimation, 'xyz' is failed in the beginning because the yaw component also appeared in $x$ component, not only $z$, due to the coincidence of the $x$ and $z$ rotation.
Our method can achieve equivalence performance, while direct fusion requires careful calculation by hand.
In addition, ours can involve additional information other than IMUs into the estimation without additional implementation.

\begin{figure}[t]
    \centering
    \includegraphics[width=\linewidth]{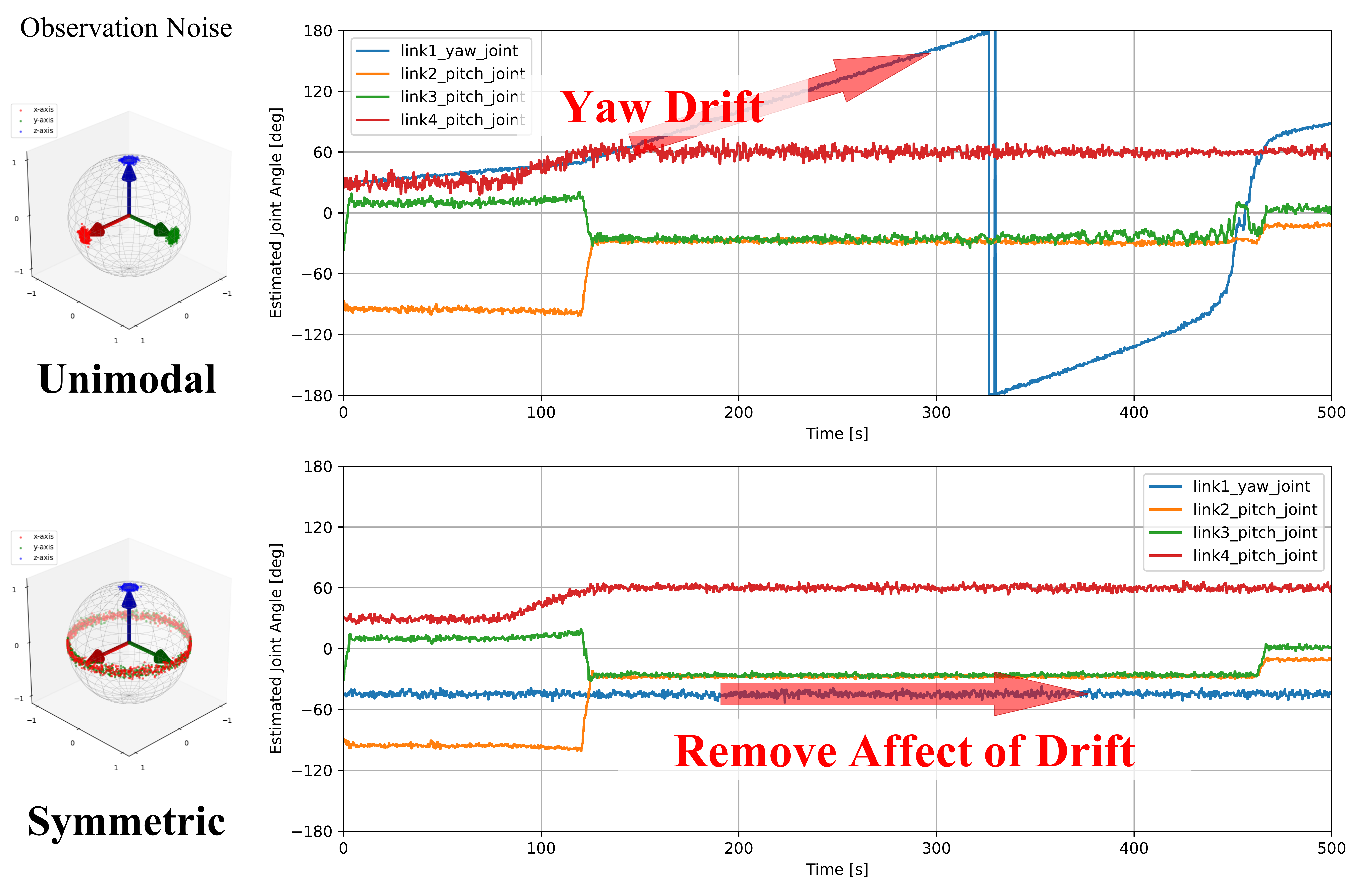}
    \vspace{-6mm}
    \caption{Comparison of the results with using unimodal and axis-symmetric distribution for observation noise of the IMUs.}
    \label{fig:driftcancel}
  \end{figure}

\label{section:IMU}

\subsection{Probabilistic Representation of IMU drift}


To see the effect of introducing directional statistical distributions, we present a method for yaw drift suppression of IMU sensors using an axisymmetric Bingham distribution.
A magnetic sensor is often used for observing the yaw direction, but it is difficult to use here because the sensors in the robot are surrounded by the equipment generating noisy electromagnetic fields.
We use the 6-axis IMU sensor to estimate the links' orientation, although it has a yaw drift.
Our method solves this drifting problem by using the axis-symmetric distribution as the observation noise.
\figref{fig:driftcancel} shows the fusion results and implies the symmetric distribution can absorb the yaw drift.
This is because likelihoods are the same with any quaternion around the $z$-axis, so the system uses other information, such as the reference angle.

Our implementation uses the Madgwick filter to estimate the yaw direction from the IMU sensor.
The yaw direction can be estimated to some extent from the information on angular velocity, which can be obtained from the gyro sensor, around the direction of gravity, but that information is also dropped.
Further discussion is needed to use this information wisely.
Instead of estimating the yaw direction from the gyro sensor, if any other information about the yaw direction is available, such as the gripper position, the effective joint angle can be estimated that is consistent with this additional information (see the fusion result of \verb|link1_yaw_joint| in column 1 and 2 of \figref{fig:simulationresult}).
\section{DISCUSSION}

\subsection{Varying Hyperparameters} \label{sec:hyperparameters}

The hyperparameters used in our experiments are shown in \tabref{tab:hyperparameters}. 
We researched the influence of the choice of parameters. 

\begin{enumerate}
    \item \textbf{Initial distribution}: It has little effect on results.
    \item \textbf{System noise}: The estimation becomes very noisy for small $\kappa$. 
    Too large $\kappa$ may lead particles to degenerate.
    \item \textbf{The number of particles}: The calculation becomes faster for smaller $N$, but too small $N$ would lead to degenerated particles. $N \geq 256$ is required for our case.
    \item \textbf{Bingham distribution's $\eigval$ for links}: The larger the Bingham parameters' magnitude is, the more concentrated the joint2's and joint3's distributions are. 
    \item \textbf{Gaussian distribution's covariance for EE}: The denser the distribution is, the denser the estimated \textit{joint1}'s distribution is. This seems to be because the end effector's $x,y$ position is determined mainly by yaw direction.
    \item \textbf{von Mises distribution's $\kappa$ for reference angles}: The larger the $\kappa$ is, the wider the 95\% confidence intervals are. The estimation itself is not heavily affected, but too small $\kappa$, say $1.0$, makes the estimation noisy.
\end{enumerate}






Finally, assuming joint angles obey Gaussian distribution, we have the same result as when we adopt von Mises distribution. However, von Mises distribution must be used when calculating the mean or variance of angles (see \eqref{eq:circluarmean}).






\subsection{Limitation of our method}

Since this method is based on a particle filter, it can be easily applied to a wide variety of robots.
On the other hand, as the characteristics of particle filters, 
the problem of degeneration becomes non-negligible if the dimension of the state becomes larger. 
To estimate much more body parameters for auto-calibration, we must solve this problem by using the knowledge of the latest filter theory.

\section{CONCLUSION}

This paper proposed a method for online estimation of self-body posture with directional statistical distributions.
It can import various sensory data into the estimation without additional implementations.
As an application of the method, we estimated the gripper position using an actual robot and evaluated the results of direct observation with an AR marker.
In the future, we will improve the method to apply to dynamic motions and auto-calibration.




\bibliographystyle{junsrt}
\bibliography{main}

\begin{thebibliography}{10}

\bibitem{GarciaCifuentes2017}
Cristina {Garcia Cifuentes}, Jan Issac, Manuel W{\"{u}}thrich, Stefan Schaal,
  and Jeannette Bohg.
\newblock {Probabilistic Articulated Real-Time Tracking for Robot
  Manipulation}.
\newblock {\em IEEE Robot. Autom. Lett.}, Vol.~2, No.~2, pp. 577--584, 2017.

\bibitem{Tao2020}
Yong Tao, Fan Ren, Youdong Chen, Tianmiao Wang, Yu~Zou, Chaoyong Chen, and Shan
  Jiang.
\newblock {A method for robotic grasping based on improved Gaussian mixture
  model}.
\newblock {\em Math. Biosci. Eng.}, Vol.~17, No.~2, pp. 1495--1510, 2020.

\bibitem{Cantelli2015}
Luciano Cantelli, Giovanni Muscato, Marco Nunnari, and Davide Spina.
\newblock {A Joint-Angle Estimation Method for Industrial Manipulators Using
  Inertial Sensors}.
\newblock {\em IEEE/ASME Trans. Mechatronics}, Vol.~20, No.~5, pp. 2486--2495,
  2015.

\bibitem{von2016human}
Timo Von~Marcard, Gerard Pons-Moll, and Bodo Rosenhahn.
\newblock Human pose estimation from video and imus.
\newblock {\em IEEE transactions on pattern analysis and machine intelligence},
  Vol.~38, No.~8, pp. 1533--1547, 2016.

\bibitem{Lee2020}
Michelle~A. Lee, Brent Yi, Roberto Martin-Martin, Silvio Savarese, and
  Jeannette Bohg.
\newblock {Multimodal sensor fusion with differentiable filters}.
\newblock {\em IEEE Int. Conf. Intell. Robot. Syst.}, pp. 10444--10451, 2020.

\bibitem{Brookshire2016}
Jonathan Brookshire and Seth Teller.
\newblock {Articulated pose estimation using tangent space approximations}.
\newblock {\em Int. J. Rob. Res.}, Vol.~35, No. 1-3, pp. 5--29, jan 2016.

\bibitem{Lu2022}
Guozheng Lu, Wei Xu, and Fu~Zhang.
\newblock On-manifold model predictive control for trajectory tracking on
  robotic systems.
\newblock {\em IEEE Transactions on Industrial Electronics}, pp. 1--10, 2022.

\bibitem{chen2015}
Chen Muyi and Wang Hongyuan.
\newblock A new recursive filter based on the gauss von mises distribution.
\newblock In {\em 2015 IEEE 6th International Symposium on Microwave, Antenna,
  Propagation, and EMC Technologies (MAPE)}, pp. 329--332, 2015.

\bibitem{CHEN2016543}
Mu~yi~Chen and Hong yuan Wang.
\newblock Nonlinear measurement update for recursive filtering based on the
  gauss von mises distribution.
\newblock {\em Procedia Computer Science}, Vol.~92, pp. 543--548, 2016.
\newblock 2nd International Conference on Intelligent Computing, Communication
  \& Convergence, ICCC 2016, 24-25 January 2016, Bhubaneswar, Odisha, India.

\bibitem{igor2014}
Igor Gilitschenski, Gerhard Kurz, Simon~J. Julier, and Uwe~D. Hanebeck.
\newblock Efficient bingham filtering based on saddlepoint approximations.
\newblock In {\em 2014 International Conference on Multisensor Fusion and
  Information Integration for Intelligent Systems (MFI)}, pp. 1--7, 2014.

\bibitem{binghamfilter2018}
Rangaprasad~Arun Srivatsan, Mengyun Xu, Nicolas Zevallos, and Howie Choset.
\newblock Probabilistic pose estimation using a bingham distribution-based
  linear filter.
\newblock {\em The International Journal of Robotics Research}, Vol.~37, No.
  13-14, pp. 1610--1631, 2018.

\bibitem{Zhou2019OnTC}
Yi~Zhou, Connelly Barnes, Jingwan Lu, Jimei Yang, and Hao Li.
\newblock On the continuity of rotation representations in neural networks.
\newblock {\em 2019 IEEE/CVF Conference on Computer Vision and Pattern
  Recognition (CVPR)}, pp. 5738--5746, 2019.

\bibitem{Mardia1999}
Kanti~V. Mardia and Peter~E. Jupp.
\newblock {\em {Directional Statistics}}.
\newblock Wiley Series in Probability and Statistics. John Wiley {\&} Sons,
  Inc., Hoboken, NJ, USA, jan 1999.

\bibitem{Hassan2012}
S.F. Hassan, A.G. Hussin, and Y.Z. Zubairi.
\newblock {Improved efficient approximation of concentration parameter and
  confidence interval for circular distribution}.
\newblock {\em ScienceAsia}, Vol.~38, No.~1, p. 118, 2012.

\bibitem{bingham1974}
Christopher Bingham.
\newblock {An Antipodally Symmetric Distribution on the Sphere}.
\newblock {\em The Annals of Statistics}, Vol.~2, No.~6, pp. 1201 -- 1225,
  1974.

\bibitem{Gordon1993}
N.J. Gordon, D.J. Salmond, and A.F.M. Smith.
\newblock {Novel approach to nonlinear/non-Gaussian Bayesian state estimation}.
\newblock {\em IEE Proc. F Radar Signal Process.}, Vol. 140, No.~2, p. 107,
  1993.

\bibitem{Li2015}
Tiancheng Li, Miodrag Bolic, and Petar~M. Djuric.
\newblock {Resampling Methods for Particle Filtering: Classification,
  implementation, and strategies}.
\newblock {\em IEEE Signal Process. Mag.}, Vol.~32, No.~3, pp. 70--86, may
  2015.

\bibitem{pinocchioweb}
Justin Carpentier, Florian Valenza, Nicolas Mansard, et~al.
\newblock Pinocchio: fast forward and inverse dynamics for poly-articulated
  systems.
\newblock https://stack-of-tasks.github.io/pinocchio, 2015--2021.

\bibitem{madgwick2010efficient}
Sebastian Madgwick, et~al.
\newblock An efficient orientation filter for inertial and inertial/magnetic
  sensor arrays.
\newblock {\em Report x-io and University of Bristol (UK)}, Vol.~25, pp.
  113--118, 2010.

\end{thebibliography}

\end{document}